\documentclass[fleqn,12pt]{wlscirep}

\usepackage{graphicx}
\usepackage{subfigure}
\usepackage{color}
\usepackage[export]{adjustbox}

\title{Towards a Crowd Analytic Framework For Crowd Management in Majid-al-Haram}

\author[1,*]{Sultan Daud Khan}
\author[]{Muhammad Tayyab}
\author[]{Muhammad Khurram Amin}
\author[]{Akram Nour}
\author[]{Anas Basalamah}
\author[]{Saleh Basalamah}
\author[1,2,*]{Sohaib Ahmad Khan}

\affil[1]{Technology Innovation Center, Wadi Makkah, Makkah Al Mukarramah, Saudi Arabia}
\affil[2]{Science and Technology Unit, Umm Al Qura University, Makkah Al Mukarramah, Saudi Arabia}
\affil[*]{sdkhan,skhan@gistic.org}

\begin{abstract}
The scared cities of Makkah Al Mukarramah and Madina Al Munawarah host millions of pilgrims every year. During Hajj, the movement of large number of people has a unique spatial and temporal constraints, which makes Hajj one of toughest challenges for crowd management. In this paper, we propose a computer vision based framework that automatically analyses video sequence and computes important measurements which include estimation of crowd density, identification of dominant patterns, detection and localization of congestion. In addition, we analyze helpful statistics of the crowd like speed, and direction, that could provide support to crowd management personnel. The framework presented in this paper indicate that new advances in computer vision and machine learning can be leveraged effectively for challenging and high density crowd management applications. However, significant customization of existing approaches is required to apply them to the challenging crowd management situations in Masjid Al Haram. Our results paint a promising picture for deployment of computer vision technologies to assist in quantitative measurement of crowd size, density and congestion.

\end{abstract}
\begin{document}

\flushbottom
\maketitle
%
%
\thispagestyle{empty}

\section{Introduction}

The percentage of Muslims in the world is expected to increase from the current 23\% to 29.7\% by 2050 \cite{pewresearch}. With this potential increase in population, there will be an increase in the number of people performing Hajj and Umrah in Makkah Al Mukarramah. To ensure public safety, it is critical to understand crowd dynamics and congestion circumstances at Hajj and Umrah locations. However, despite significant advances in crowd management, safety measures and adoption of technology, crowd related events and disasters can still occur. Due to the complex dynamics of the crowd, crowd management has become a daunting job requiring huge efforts from the security staff to manage potentially problematic situations.


In high density crowded areas, surveillance cameras are generally installed in different locations to provide video coverage to security personnel. The current practice of understanding crowd behavior and pedestrian dynamics for the prevention of crowd disasters is often based on manual reviewing of video data and employing simulation models to identify potentially disastrous areas. Based on rules of thumb gained from previous experience as well as simulation models \cite{still2014introduction}, limits on the number of people that can be simultaneously present at a venue are decided. Such studies are also used to suggest structural changes, and to design new venues. For example, the study of Helbing et al. in~\cite{helbing2007dynamics} led to the redesign of Jamarat Bridge in Makkah Al Mukarramah in 2007. 

Several physicals models for understanding pedestrian dynamics have been proposed which are based on the analogies to gas and fluid dynamics to account for individual behaviour. The popular among them are the \emph{Social force model} in~\cite{helbing1995social} and \emph{Cellular automata} in~\cite{burstedde295simulation} which model pedestrian dynamics on a microscopic level. In~\cite{helbing2000simulating}, the basic social force model extended in order to incorporate effects of panic by adding further random forces. Experimental studies are conducted in~\cite{helbing2007dynamics,steffen2010methods} in order to understand human behaviour and improve the existing physical models by incorporating more parameters such as crowd density, speed, flow, and crowd pressure. One of the limitations of using physical models is, the results and observations are based on experimental data, captured in a controlled environment, and do not consider real-time data. These physical models require calibration and validation by means of real-time data. 

An alternative approach is to use surveillance video data to automatically monitor crowd motions, crowd densities and identify potential choke points in real-time. There is an increased interest in scientific research on automated crowd analysis in recent years \cite{li2015crowded}. In automated video surveillance, detection and tracking of objects are the core preprocessing steps. This technology requires low-level motion and appearance features~\cite{stauffer2000learning,viola2005detecting}, works well in low density situations but often fails in high density situations due to less pixels per person, severe occlusions and clutter. Therefore, the researchers proposed \emph{holistic approaches}~\cite{khan2016analyzing} by gathering global motion information at a higher level which is captured by computing optical flow, a low-level local motion feature. This approach performs well in different scenarios of crowd and it does not involve detection and tracking.

\begin{figure}[t]
\includegraphics[width=1\textwidth]{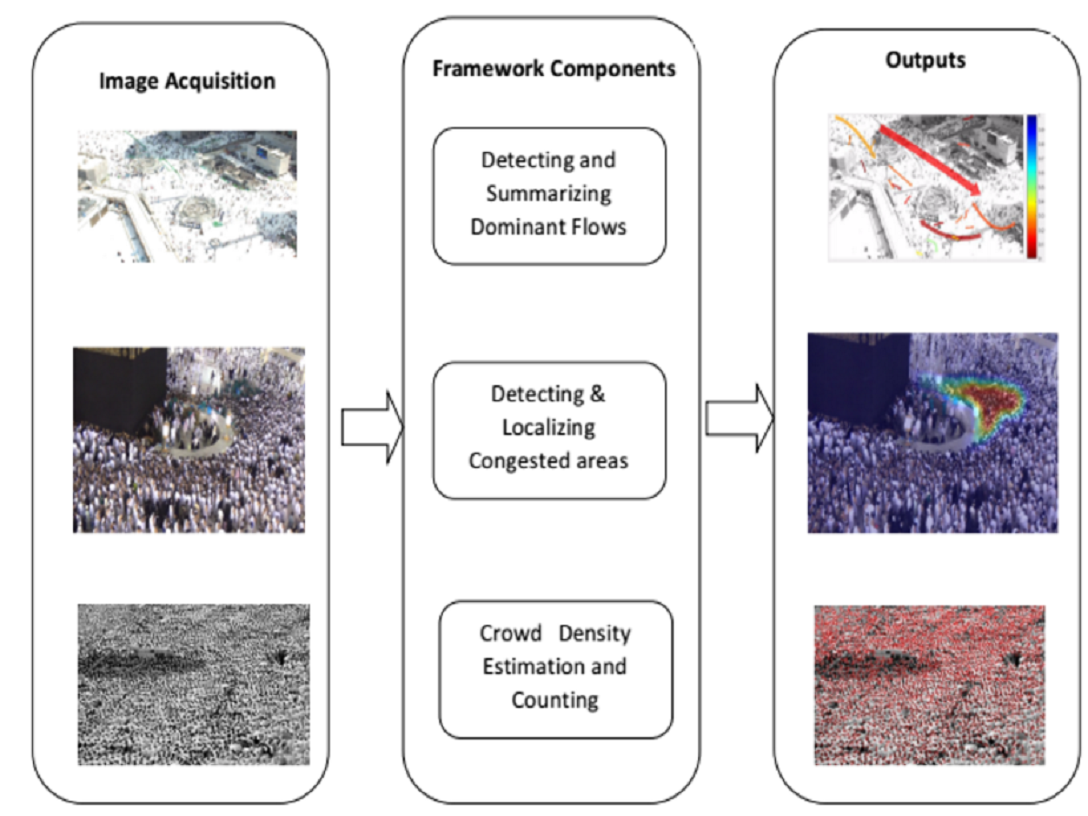}
\centering
\caption{Crowd Analytic Framework.}
   \label{fig:framework}
\end{figure}

In this paper, we propose a computer vision based crowd analytic framework that automatically analyses video streams from surveillance IP cameras installed at different location of the scene.  Using IP camera as a measuring device, the system computes important measurement which include estimation of crowd density, identification and summarization of dominant motion patterns, and detection and localization of congested areas, that could provide support to crowd management personnel. With this surveillance setup, security staff deployed for public safety and security can efficiently and effectively respond to the evolving anomalous and potentially dangerous situations. Our proposed system is based on three major modules. An overview of our proposed framework is shown in Figure~\ref{fig:framework}. 

The first module of our proposed framework will \textit{identify and summarize dominant flows} of a crowd in a simplified output that can aid surveillance operators and crowd management personnel quickly understand the dynamics of the scene in a single glance. This summarized view tackles the well-researched problem of fatigue in surveillance operators, which has been shown to limit the effectiveness of surveillance systems significantly. Surveillance personnel looking at a video wall with multiple screens, showing slowly evolving situation have been shown to easily lose concentration and miss out important information. By summarizing the scene in a few dominant flows, the situation can be reviewed easily and effectively. Our proposed system can automatically create these concise scene summaries by computing the motion in multiple video scenes. 

The second module will \textit{detect and localize congested areas} by analyzing the flow of the crowd, and identify locations where crowd flow is obstructed. An efficient implementation of this system can substantially reduce the cost by deploying exact number of security personnel and reduce causalities to a great extent by responding swiftly to the anomalous situations. We have developed this novel congestion detection and localization algorithm that can work on any type of video feed and can identify congestion areas in real-time. 

The third module is to \textit{estimate crowd density}. It will detect and localize the individuals in the crowd to generate an overall count and density in any scene. Accurate crowd density estimator is crucial in situations where critical crowd management actions are predicated on measuring the density of the current crowd. For example, the ingress of more pilgrims in Mataaf may have to be restricted if the density or the count within Mataaf crosses a certain threshold. Similarly, certain precautionary measures may have to be initiated in Mina if the density of pilgrims on the roads is observed to be excessive. Measurement of accurate crowd statistics may also help in deployment of optimal number of security personnel.

Management thinker Peter Drucker is often quoted as saying, ``if you can't measure it, you can't improve it''. This is especially true for the crowd management situation faced in Masjid Al Haram and Masha'er Al Muqaddasah. While the authorities have, over the years, developed impressive expertise in managing the pilgrims, based on rules of thumb and field experience, it is still a worthy goal to move towards accurate modeling of the critical crowd dynamics situations that arise during the pilgrimage. Measurement and identification of issues is the first and most critical step in this direction, which will enable new paradigms in crowd management and ensure safe and secure experience for the pilgrims.  

In the following section, we briefly discuss each core component of our framework.

\section{Framework Components}

In this section, we briefly discuss the proposed framework components

\subsection{Geometric Rectification}

As a preprocessing step our system creates a rectified panorama view from multiple sources of overlapping images or videos. Here we briefly discuss this preprocessing step. \\

\begin{figure}[t]
\centering
\subfigure[Geometric Rectification creates a view of the plane as if the camera was at the top.]{\label{fig:rectifcation}\includegraphics[width=0.8\textwidth]{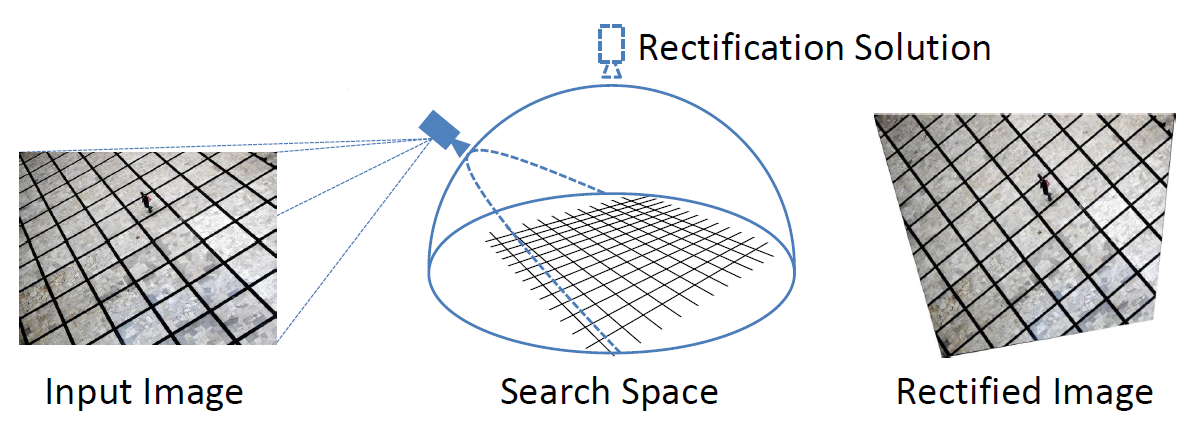}}\\
\subfigure[Panorama view created from multiple images.]{\label{fig:pano}\includegraphics[width=0.8\textwidth]{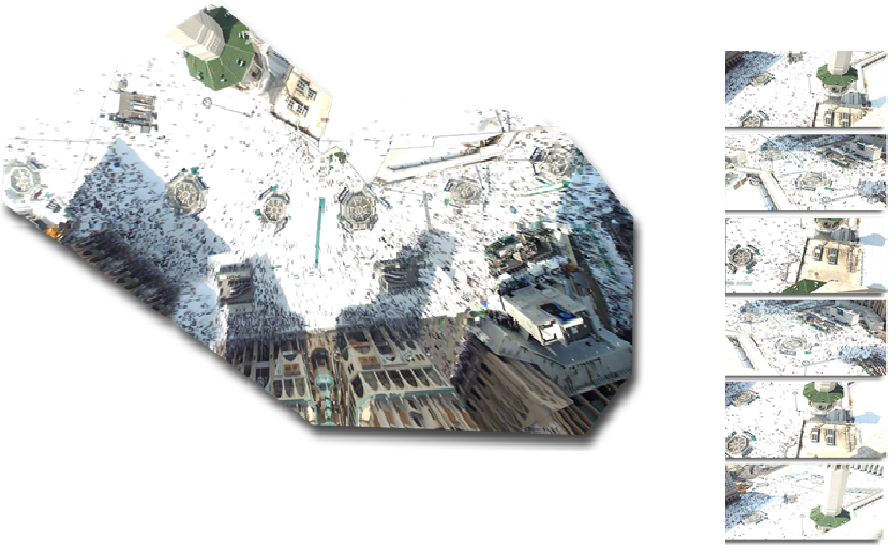}}
\caption{Geometric Rectification and Panorama Generation}
   \label{fig:preprocess}
\end{figure}

Pinhole cameras capture the world in a way that creates perspective foreshortening effect. This effect distorts the world measurements in a way that the distances closer to the camera appear larger than those far away even when they are same (see {Figure~\ref{fig:rectifcation}). While humans are capable of correcting this distortion in our perception, computers have no such training and this distortion affects their computation. This distortion can be corrected using a the geometric model of a pinhole camera, through a process know as rectification. \\
Rectification has an added advantage that it also enables us to fuse images and videos from multiple cameras into single panoramic view as illustrated in {Figure~\ref{fig:pano}. This is possible because after rectification, all the images share a common virtual view-point, as the scene would have appeared like from the top. Thus they can be fused together into a larger view spanning all the cameras.

\subsection{Detection and Summarization of Dominant Flows}

An important contribution that automated analysis tools can generate for management of pedestrians and crowd safety is the detection of conflicting large pedestrian flows: this kind of movement pattern, in fact, may lead to dangerous situations and potential threats to pedestrians' safety. Therefore, segmenting and summarizing typical flow patterns of crowd are important steps to understand overall crowd dynamics. 

For detecting dominant motion patterns, we need to extract motion information from the scene. Trajectories of pedestrians capture the local motion information of the video. Long and dense trajectories, that is, trajectories representing a large number of paths followed by different pedestrians, reaching a significant length, provide good coverage of foreground motion as well as of the surrounding context. In high density situations, detection and tracking of pedestrians is not reliable due to few pixels per individual, clutter and extreme occlusions. Therefore, in order to get rid of the limitations of traditional methods, we adopt \emph{Particle Advection} approach which capture global motion information from the scene and is based on optical flow computation. The approach for detecting dominant flows starts by dividing the input video into multiple segments of equal length and duration, considering videos with a constant frame rate. The initial frame of each segment is overlaid by a grid of particles initializing a dynamical system defined by optical flow. Time integration of the dynamical system over a segment of the video provides particle trajectories (tracklets) that represent motion patterns in the scene for a certain time interval associated to the analyzed segment. We detect dominant flows by clustering of tracklets, obtained using an unsupervised hierarchical clustering algorithm, where the similarity is measured by the Longest Common Sub-sequence (LCS) metric. Motions patterns in the scene are summarized and represented by using color coded arrows, where speeds of the different flows are encoded with colors, the width of an arrow represents the density (number of people belonging to a particular motion pattern) while the arrow head represents the direction. This novel representation of crowded scene provides a clutter free visualization which helps the crowd managers in understanding the scene. The detailed description of the approach can be found in~\cite{khan2016analyzing}.

\subsection{Detection and Localization of Congested Locations}

Crowd congestion is a condition when individuals in a crowd are prevented from moving smoothly due to over-crowding, and are unable to make progress towards their desired goal at normal speed. Congestion is characterized by involuntary slowing down or even coming to a complete standstill, long queues and waiting times. In an uncongested state, individuals move smoothly towards the destination with comfortable speeds. This intended speed and direction of pedestrians is affected by the neighboring pedestrians. As the density increases, the distance between the individuals decreases, thus affecting each others behavior and as a result individuals cease to move with their desired velocities. 

Our framework for congestion detection and localization use only easily measurable motion features to reliably localize pedestrian congestion in surveillance videos in a variety of scenarios. It has been reported in crowd dynamics literature that individuals in a crowd tend to undergo lateral oscillations, that is, to and fro motion orthogonal to their desired direction of motion, when in a congestion state \cite{liu2009extraction}. This is attributed to the shifting of weight from one foot to another, termed lateral oscillation, as longitudinal progress is restricted 
\cite{hoogendoorn2005pedestrian}. We exploit this feature to look for groups of trajectories that change from smooth motion to oscillatory harmonics, as this corresponds to a change from unrestricted flow to a substantial restriction in the ability to make progress towards the desired goal.

Our approach starts by dividing the input video into multiple overlapping temporal segments of equal duration. We extract motion information (trajectories) from each segment by particle advection approach. After extracting trajectories from each segment, we then compute oscillations in trajectories,  and generate corresponding oscillation maps. After quantizing oscillation map, we identify potential critical locations which are the possible candidate locations of congestion. By spatially and temporally exploiting the oscillation map, locations with high confidence are selected as the exact congestion locations.

\subsection{Crowd Counting and Density Estimation}
The pipeline of our counting method can be broken down into two components. First component is a convolutional neural network that acts as a head detector followed by the second component that is non-maxima suppression. \\
We started from a pre-trained head detector provided by \cite{vu15heads}. They trained this network on images of human faces extracted from Hollywood movies. In its original form, however, this head detector is of limited use to us since we are working with very high density images and in high density images human head barely spans a few pixels and face is almost never recognizable. Therefore we fine-tuned this network on our own training images. For training we extracted perspective-aware patches that only contain one head an feed them to the network as positive sample. Negative sample were generated from the background and visible human torso, since we want our head detector to give high response only on heads and not on other body parts. \\
Since our head detector is patch based, so we we need to perform one more to step before we can proceed to localize people in an image. This step involves taking network prediction on every pixel in the image. In practice we take a perspective-aware patch centered at every 3rd pixel in the image to save computation time. The result of this step is response matrix whose both dimensions are one third of the image dimensions. We resize that matrix to same dimensions as image using standard binomial interpolation. \\
Finally for localization we employed a non-maxima suppression technique inspired by \cite{rodriguez11b}. Without using density as prior we solved an optimization problem that minimizes the least squared difference between the raw response matrix and a synthetic response matrix. We generate the synthetic response matrix by convolving the average response of head detector on positive patches with the candidate locations.

\section{Experimental Results and Discussions}

This section discusses the qualitative analyses of the results obtained from experiments. We carried out our experiments on a PC of 2.6 GHz (Core i7) with 16.0 GB memory and video sequences obtained from other research groups and acquired through field observations. The videos have different field of view, resolution, frame rates and duration, yet our method performed well in all cases. We first divide each video into a set of temporal segments with 25\% overlap. While developing our framework, we solve three different problems, therefore we evaluate and report the results of each method separately.

\subsection{Crowd Counting and Density Estimation}
We tested our counting method on hight density images of Mataaf. Figure~\ref{fig:crowdlocalization} is the result of our localization method. Red dots in the image represent detections output by our system. It can be seen that these detections are very accurate which is also evident from the count. For this particular image we have manually counted the number of persons visible in the Mataaf area to be 2961 people while our system detects 2661, which is an accuracy of 90\%.

Since our system detects the location of every person in the image we can map this information into a a more useful metric, that is, the number of people per square meter, as shown in Figure~\ref{fig:density}. Each colored box in this image is one meter square on the Mataaf floor and the number of people in that are represented by the color.

\begin{figure}[h]
\centering
\subfigure[Localization of individuals in a crowd. Each red dot in the image is the output of our algorithm, indicating one individual.]{\label{fig:crowdlocalization}\includegraphics[width=0.6\textwidth]{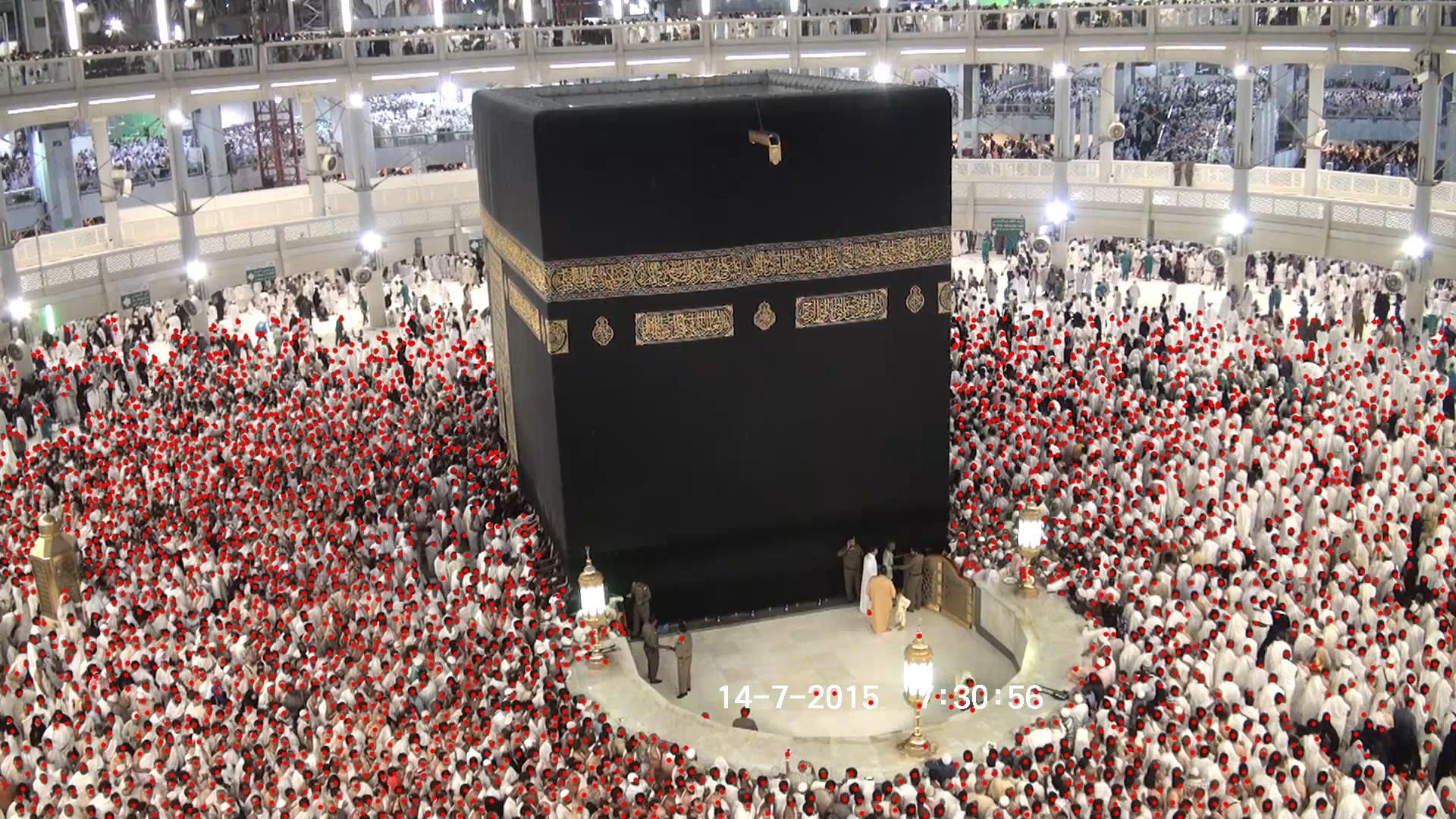}}\\

\subfigure[Density estimation in Mataaf, on persons-per-square-meter scale.]{\label{fig:density}\includegraphics[width=0.6\textwidth]{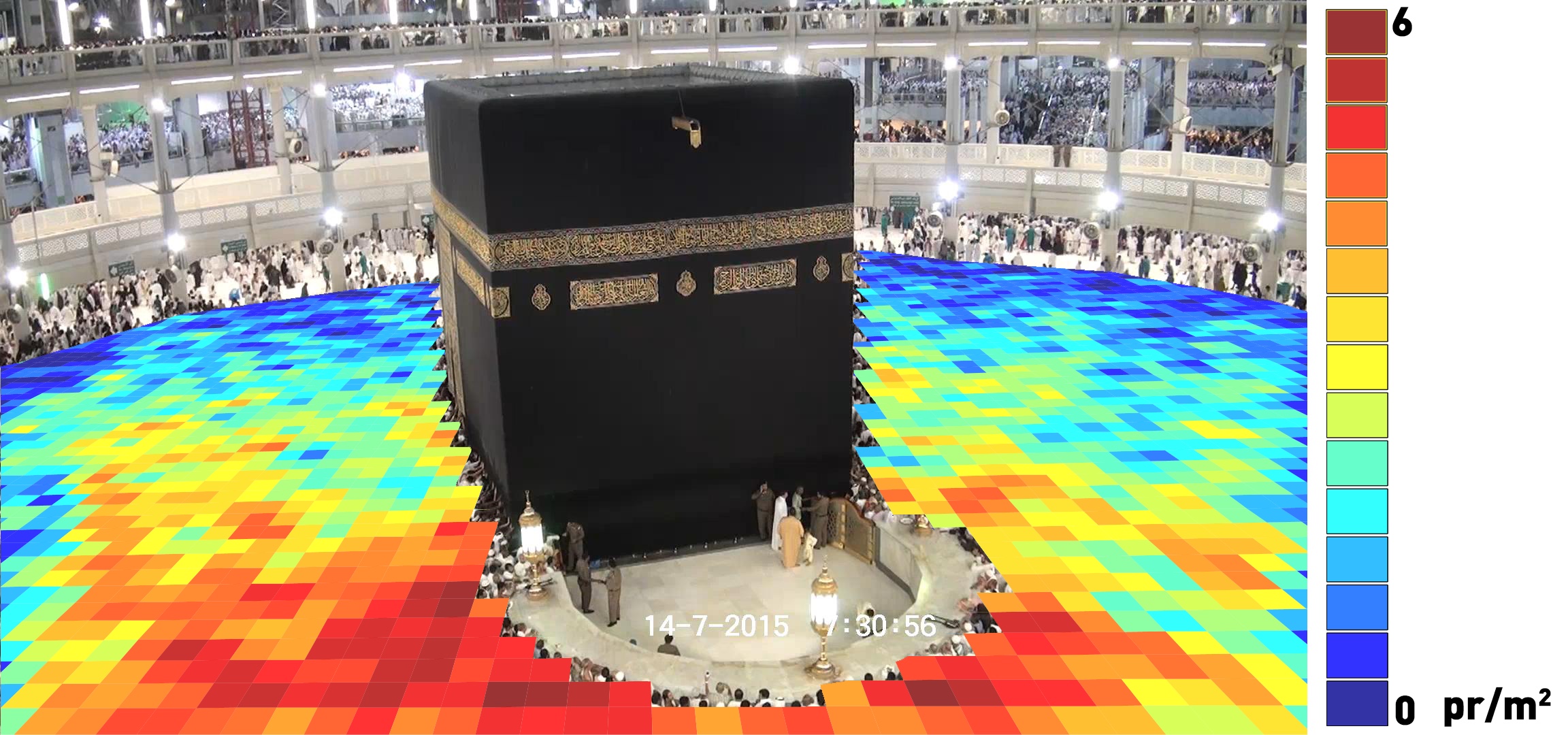}}
\caption{Crowd localization and density estimation results.}
\end{figure}

\subsection{Detection and summarization of Dominant Flow}

In order to qualitatively evaluate our framework, we used different videos covering both crowds of pedestrians and vehicles.

\begin{figure}[h]
\centering
\subfigure[Sahat video sequence during night.]{\label{fig:sahat1-1}\includegraphics[width=0.47\textwidth]{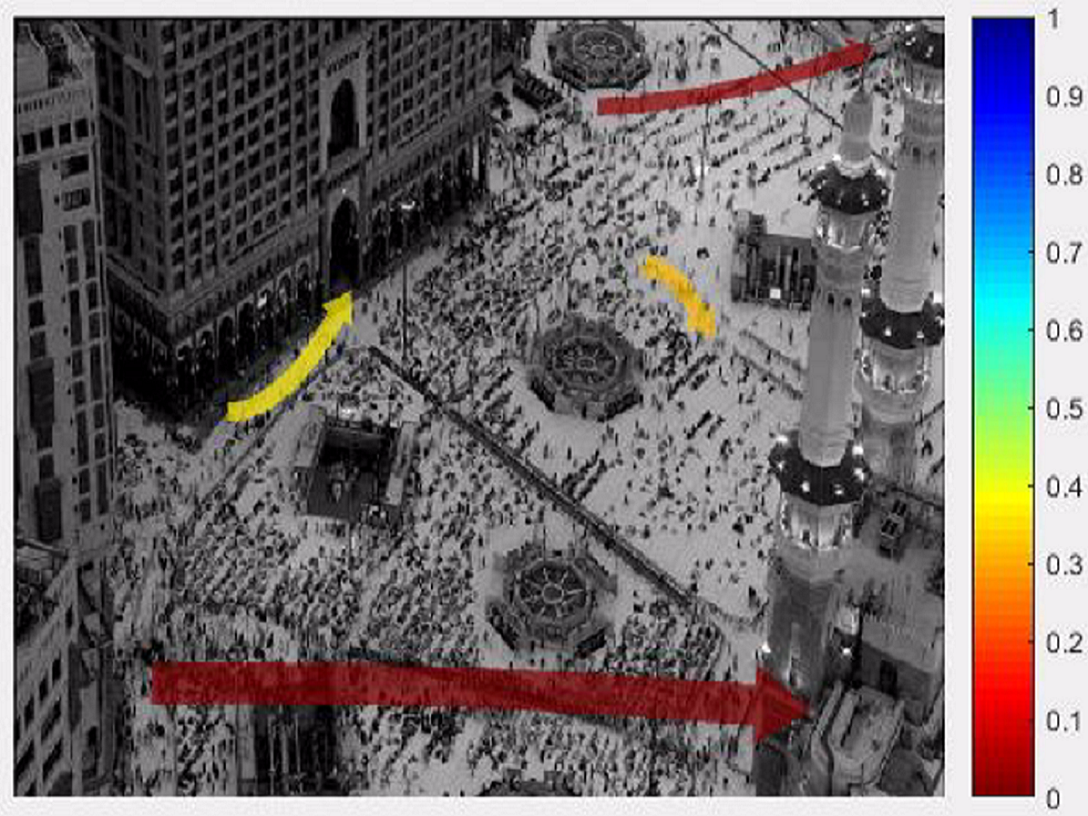}}
\subfigure[Sahat video sequence during day.]{\label{fig:sahat2-10}\includegraphics[width=0.47\textwidth]{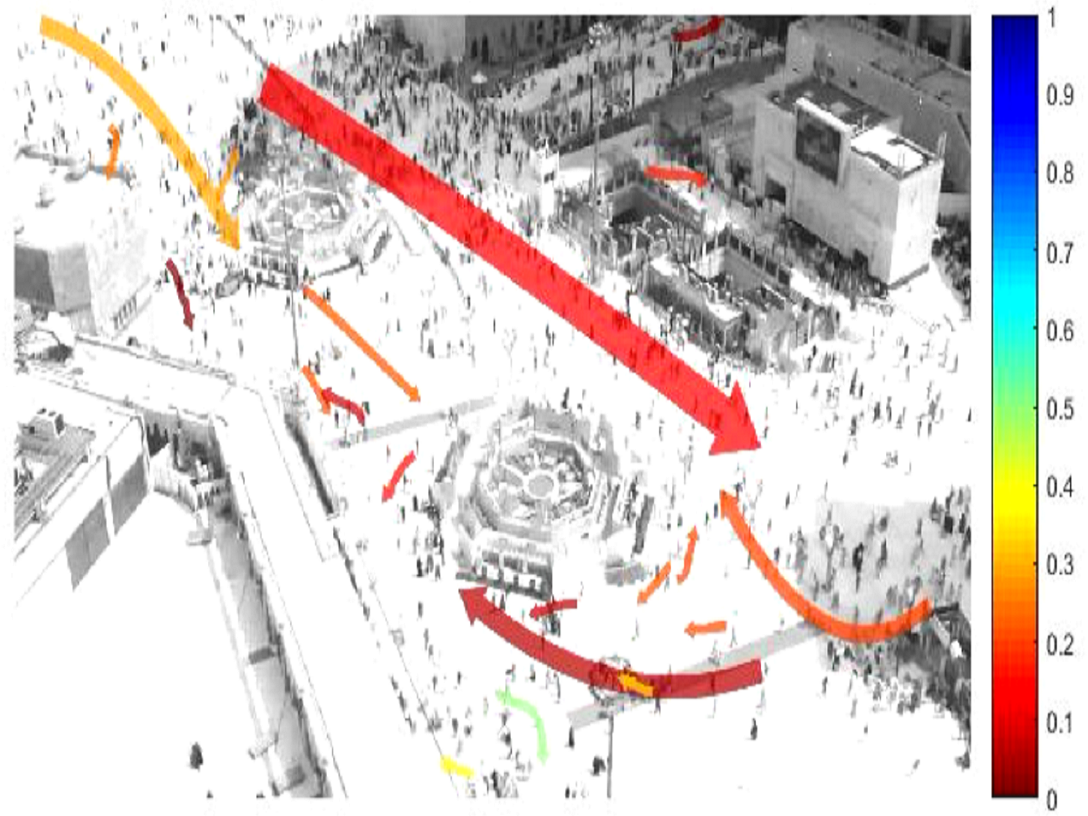}}\\
\subfigure[Traffic video sequence.]{\label{fig:traffic-20}\includegraphics[width=0.47\textwidth]{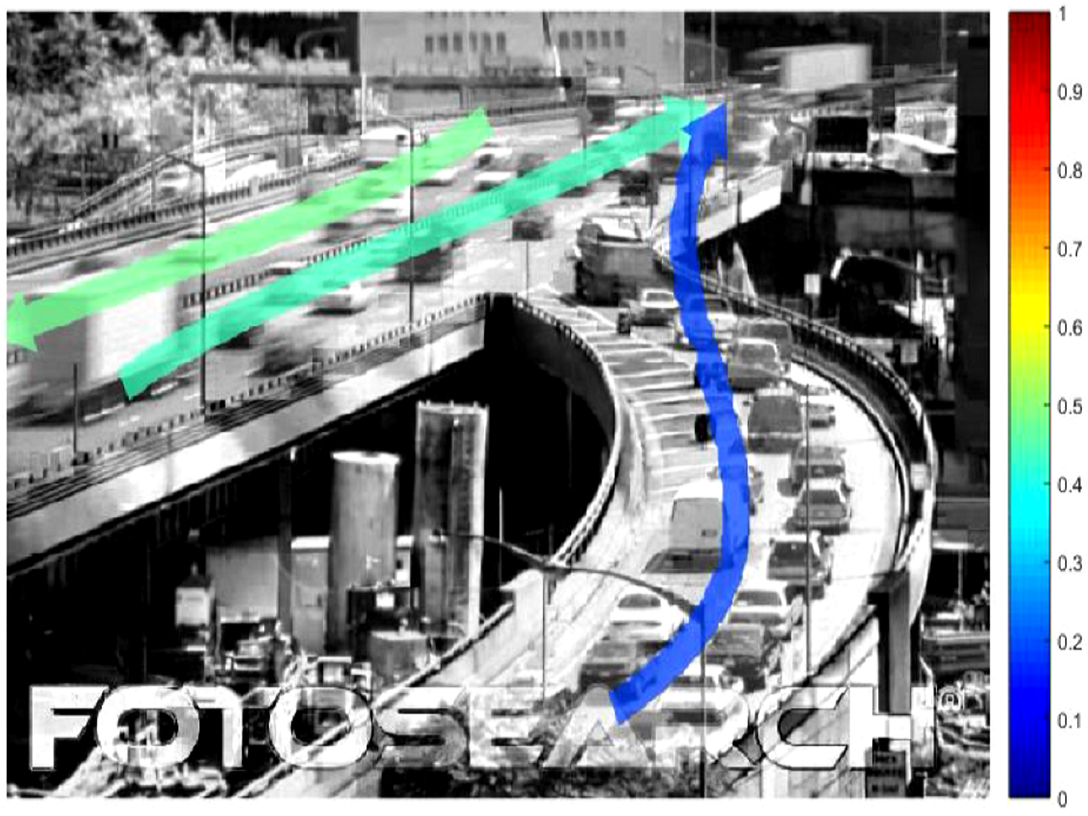}}
\subfigure[Airport video sequence.]{\label{fig:airport-50}\includegraphics[width=0.47\textwidth]{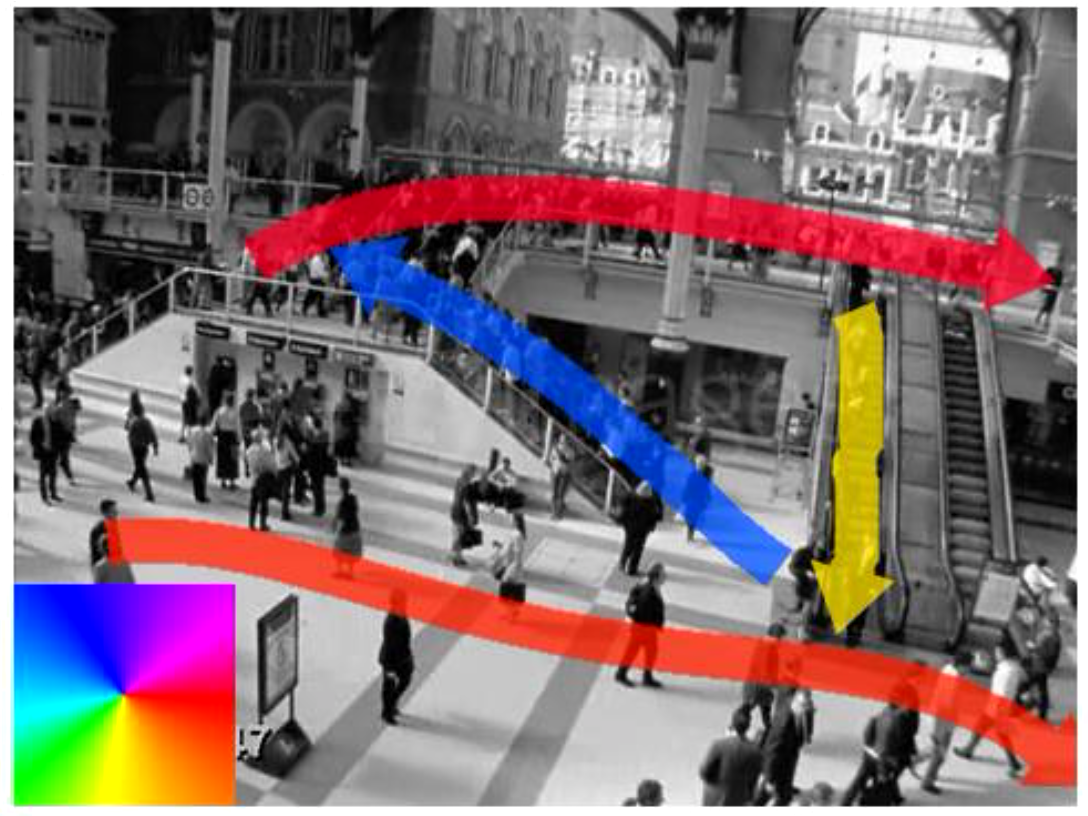}}
\caption{Summarized representation of dominant flows after the application of the algorithm.}
   \label{fig:summflows}
\end{figure}

Figure~\ref{fig:summflows} shows different videos used in our experiment. Figure~\ref{fig:sahat1-1} shows a video sequence taken from camera at height and covers saahat area  of Masjid-al-Haram where pilgrims are entering Majid-al-Haram for Isha prayers. Usually at prayers time, large number of people entering Masjid-al-Haram and this fact is illustrated in the Figure~\ref{fig:sahat1-1}, where long, wide and brown arrow represents the people entering the Haram. The width of an arrow represents the density, and speed is encoded in brown color. This analysis also validated the fact that people walk slowly in high density situations. Figure~\ref{fig:sahat2-10} represents the same situations but the video is recorded in the morning. In this video sequence, our algorithm accurately detects multiple flows in the scene. Our algorithm also accurately detects motion patterns in traffic video sequence as shown in Figure~\ref{fig:traffic-20}, where vehicles are moving in two opposite direction along the separate lanes while third lanes is merging with first one. We detect multiple motion patterns in the airport sequence as shown in Figure~\ref{fig:airport-50}. From the experimental results, it is obvious that our detection and summarization of dominant flows gives a clutter free visualization which helps the crowd managers in understanding the scene.

\subsection{Detection and Localization of Congested Locations}

In this section we evaluate qualitatively the results of our congestion detector algorithm. This dataset composed of video sequences taken from multiple cameras at Haram with different view points. These video sequences were taken in the context of the yearly pilgrimage to Makkah, Saudi Arabia, and it shows a very high density situation. The resolution of video is 1080 x 1440 pixels and composed of 28000 frames (18.66 min) long with frame rate of 25 fps.

\begin{figure}[h!]
\centering
 \subfigure[concert video sequence.]{\label{fig:concert}\includegraphics[width=0.47\textwidth]{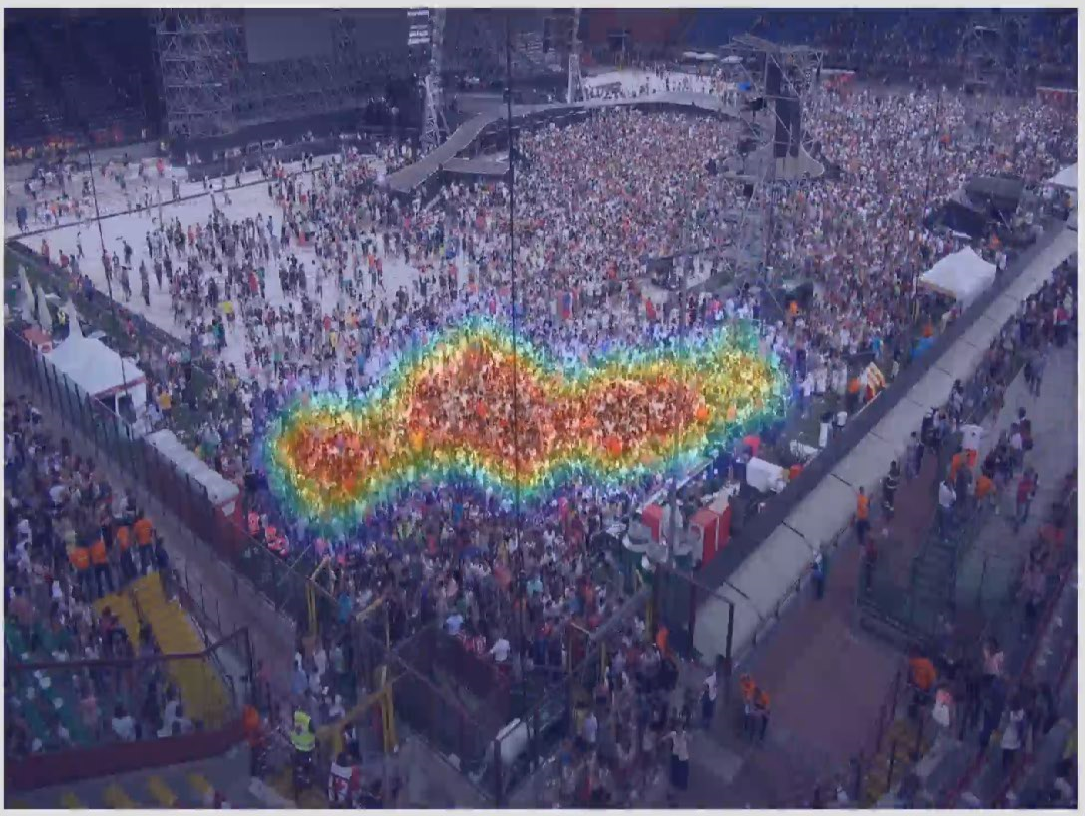}}
\subfigure[Haram1 video sequence.]{\label{fig:haram1}\includegraphics[width=0.47\textwidth]{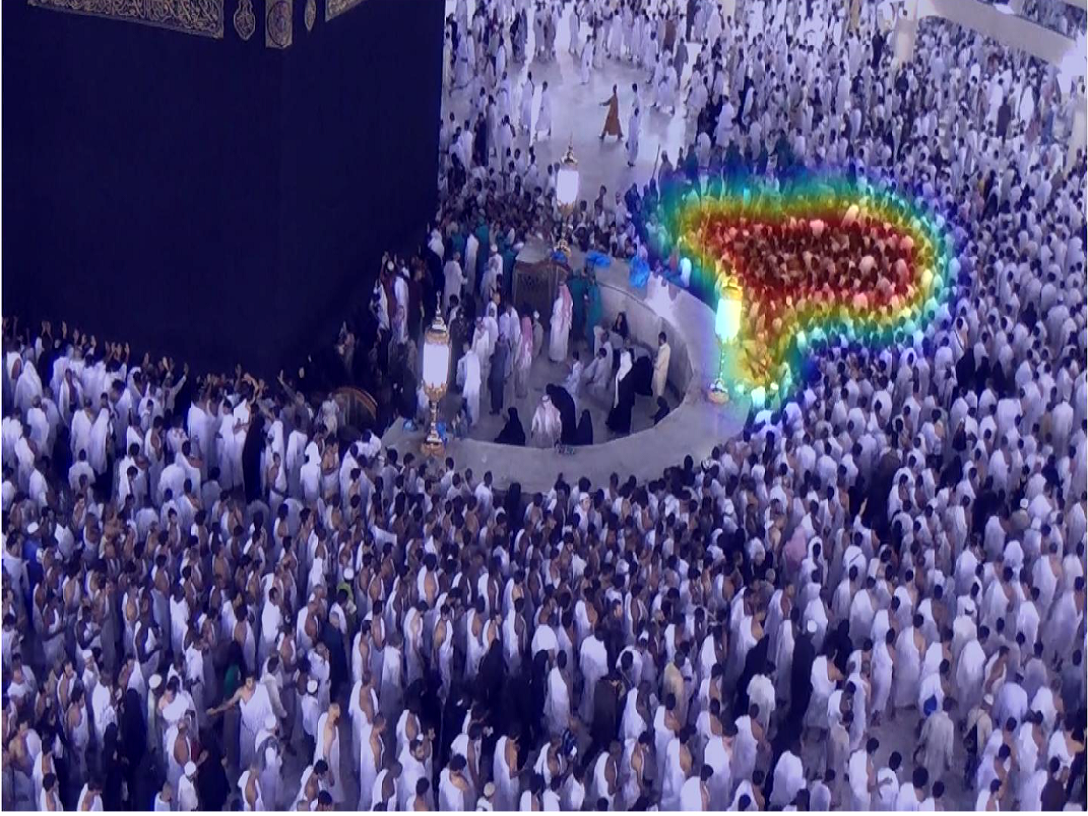}}\\
\subfigure[Haram2 video sequence.]{\label{fig:haram2}\includegraphics[width=0.47\textwidth]{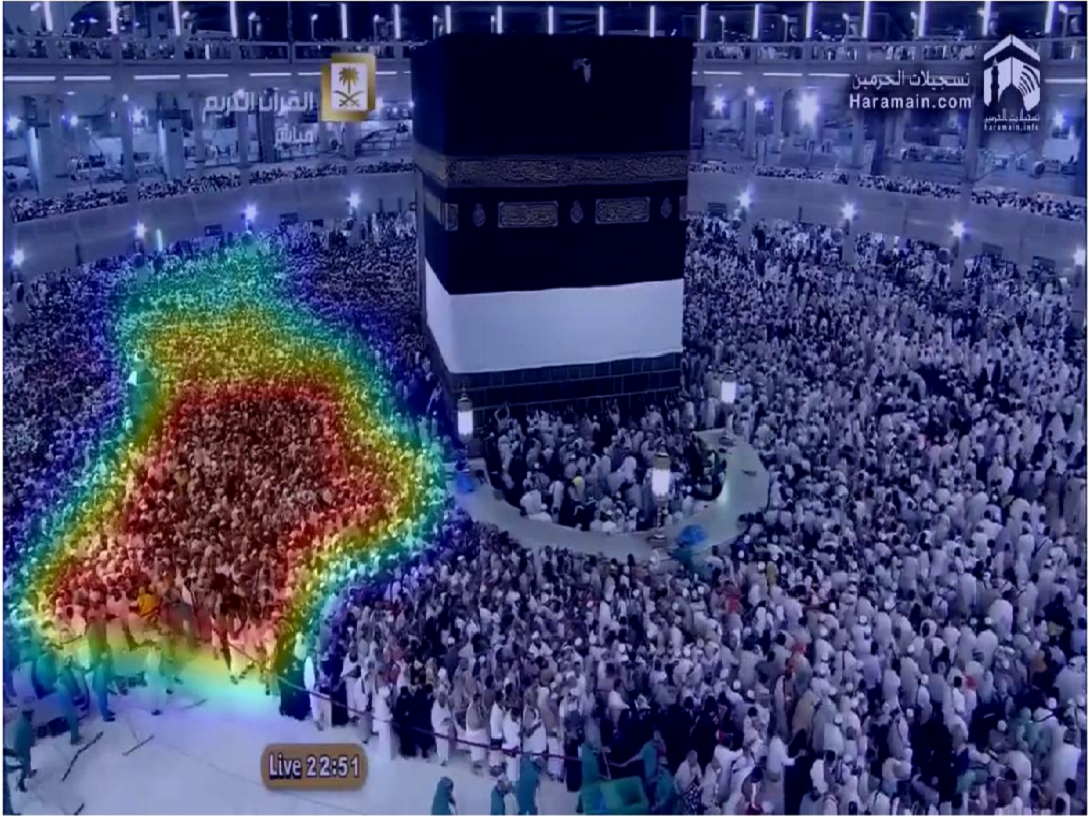}}
\subfigure[Haram3 video sequence.]{\label{fig:haram3}\includegraphics[width=0.47\textwidth]{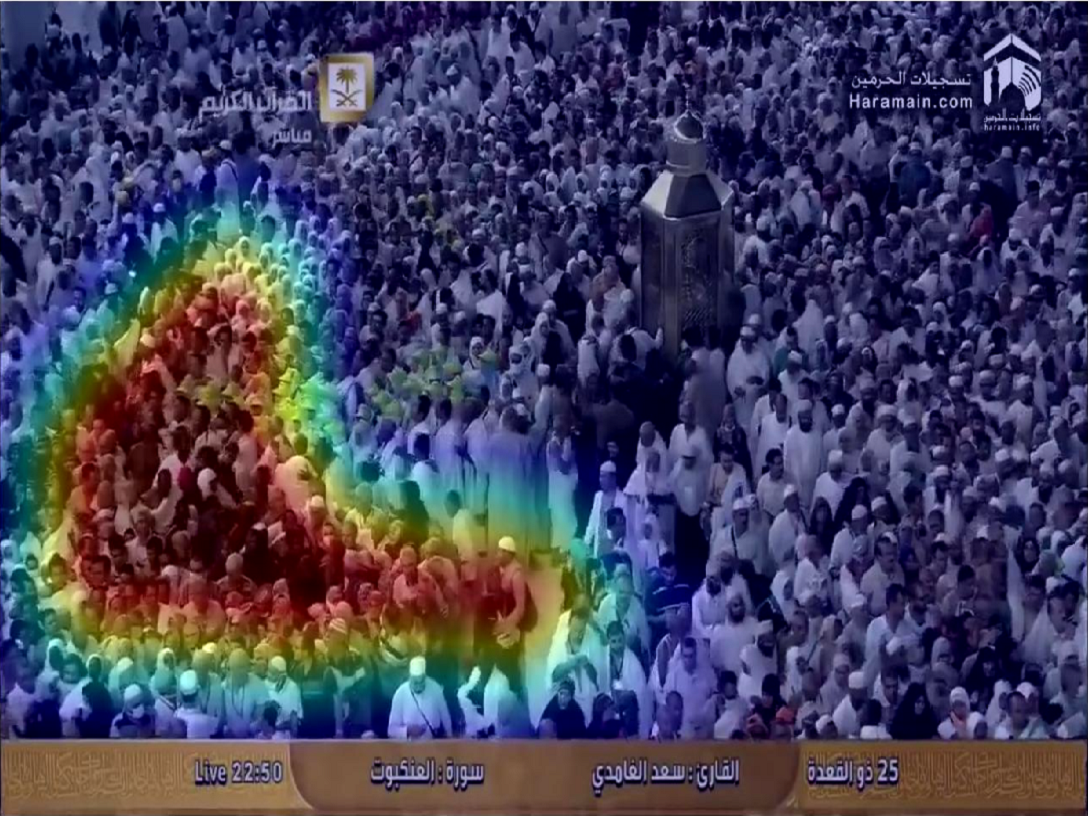}}
\subfigure[Haram4 video sequence.]{\label{fig:haram4}\includegraphics[width=0.47\textwidth]{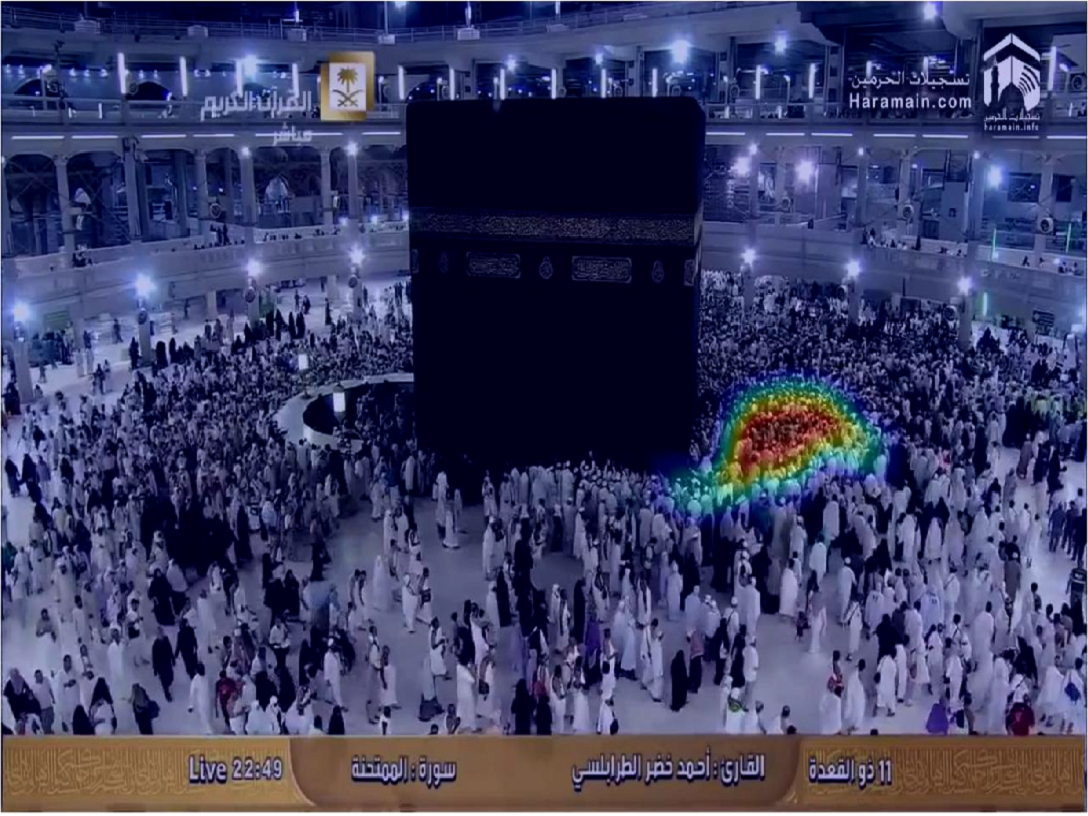}}
\subfigure[Haram5 video sequence.]{\label{fig:harm5}\includegraphics[width=0.47\textwidth]{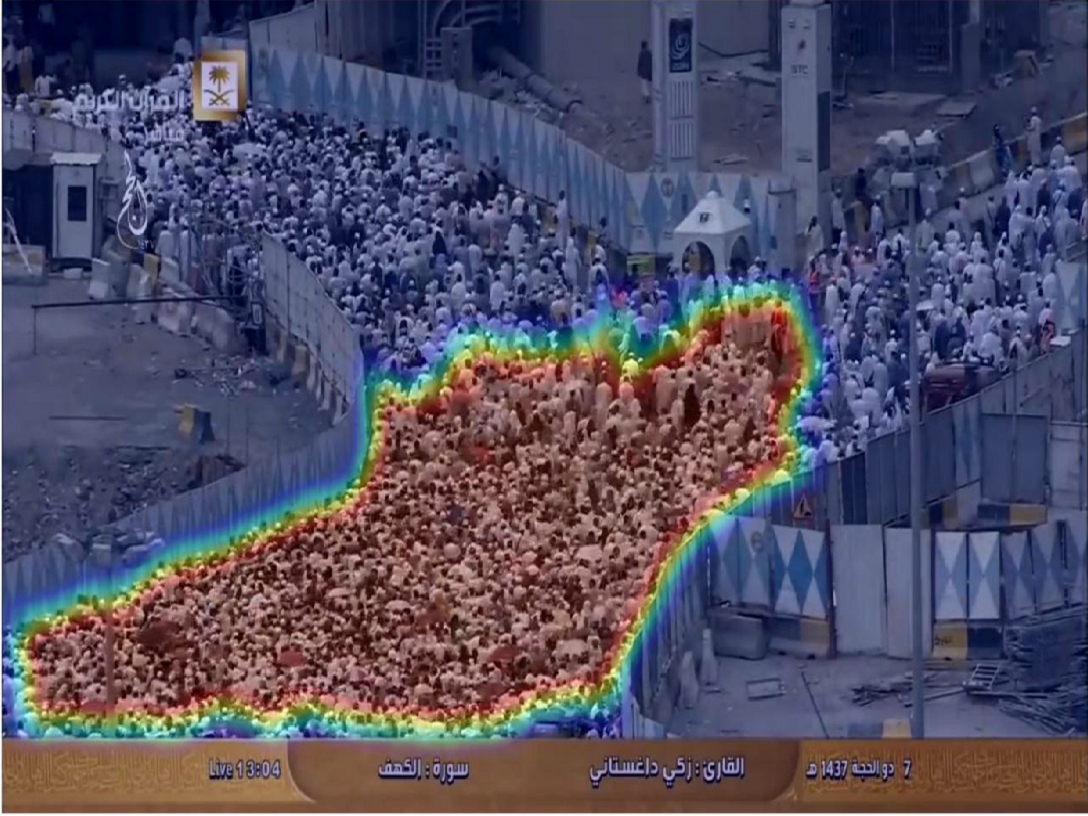}}\\

\caption{Detection and localization of congested locations after the application of our algorithm.}
   \label{fig:summflows}
\end{figure}

{Figure~\ref{fig:haram1} and Figure~\ref{fig:haram2} shows video sequences where large group of pedestrians are performing annual ritual by circulating around the kaaba. Another small group of pedestrians (the direction of which is orthogonal to large flow) trying to penetrate inside the large flow,  blocking the motion of the large flow. As the small group of pedestrians penetrates forward, causes a congestions at different locations at different temporal segments. This video sequence shows the example of \emph{Dynamic Congestion}, where the location of congested area changes with time. Figure~\ref{fig:haram3} shows a situation where a large group of people block the passage for other pilgrims while leaving the Mataf area. Similar phenomena is observed in Figure~\ref{fig:haram4}, where some pilgrims stop near Hijra-Aswad which blocks the passage for other pilgrims. In this particular low density scenario, the scale of congestion is relatively low which is obvious from the small size of the blob. Figure~\ref{fig:harm5} shows extremely high density situation where pedestrian leave the Haram area along two separate paths. This video sequence is a special case of evaluation scenario, where large number of pedestrians, after prayers, trying to leave through different paths. Similar situation is observed in Figure~\ref{fig:concert},where large number of pedestrians, after attending the concert trying to leave through a narrow exit. Intuitively, such narrow exit creates congestion which is efficiently detected by our algorithm. This is a scenario of a \emph{Fixed Congestion}, where congestion of almost the same area is created at the same locations in all temporal segments.

Figure~\ref{fig:haram1} and Figure~\ref{fig:haram2} shows video sequences where large group of pedestrians are performing annual ritual by circulating around the kaaba. Another small group of pedestrians (the direction of which is orthogonal to large flow) trying to penetrate inside the large flow,  blocking the motion of the large flow. As the small group of pedestrians penetrates forward, causes a congestions at different locations at different temporal segments. This video sequence shows the example of \emph{Dynamic Congestion}, where the location of congested area changes with time. Figure~\ref{fig:haram3} shows a situation where a large group of people block the passage for other pilgrims while leaving the Mataf area. Similar phenomena is observed in Figure~\ref{fig:haram4}, where some pilgrims stop near Hijra-Aswad which blocks the passage for other pilgrims. In this particular low density scenario, the scale of congestion is relatively low which is obvious from the small size of the blob. Figure~\ref{fig:harm5} shows extremely high density situation where pedestrian leave the Haram area along two separate paths. This video sequence is a special case of evaluation scenario, where large number of pedestrians, after prayers, trying to leave through different paths. Similar situation is observed in Figure~\ref{fig:concert},where large number of pedestrians, after attending the concert trying to leave through a narrow exit. Intuitively, such narrow exit creates congestion which is efficiently detected by our algorithm. This is a scenario of a \emph{Fixed Congestion}, where congestion of almost the same area is created at the same locations in all temporal segments.

\section{Conclusion}

We have shown in this paper that recent advances in computer vision can be leveraged to design customized algorithms for the crowd management in Masjid Al Haraam. Towards this end, we have shown that crowd counting, density and congestion can be quantitatively measured from images and video of appropriately located cameras. We also demonstrate that crowd motion information can be summarized and presented in an easily interpretable manner to the security personnel. These types of algorithms can go a long way towards helping manage pilgrims better, and ensuring their safety.




 





\bibliography{sample}

\begin{thebibliography}{10}
\expandafter\ifx\csname url\endcsname\relax
  \def\url#1{\texttt{#1}}\fi
\expandafter\ifx\csname urlprefix\endcsname\relax\def\urlprefix{URL }\fi
\expandafter\ifx\csname doiprefix\endcsname\relax\def\doiprefix{DOI }\fi
\providecommand{\bibinfo}[2]{#2}
\providecommand{\eprint}[2][]{\url{#2}}

\bibitem{pewresearch}
 \emph{\bibinfo{title}{The Future of World Religions: Population Growth
  Projections}} (\bibinfo{year}{Pew Research Center, 2015}).

\bibitem{still2014introduction}
\bibinfo{author}{Still, G.~K.}
\newblock \emph{\bibinfo{title}{Introduction to crowd science}}
  (\bibinfo{publisher}{CRC Press}, \bibinfo{year}{2014}).

\bibitem{helbing2007dynamics}
\bibinfo{author}{Helbing, D.}, \bibinfo{author}{Johansson, A.} \&
  \bibinfo{author}{Al-Abideen, H.~Z.}
\newblock \bibinfo{title}{Dynamics of crowd disasters: An empirical study}.
\newblock \emph{\bibinfo{journal}{Physical review E}}
  \textbf{\bibinfo{volume}{75}}, \bibinfo{pages}{046109}
  (\bibinfo{year}{2007}).

\bibitem{helbing1995social}
\bibinfo{author}{Helbing, D.} \& \bibinfo{author}{Molnar, P.}
\newblock \bibinfo{title}{Social force model for pedestrian dynamics}.
\newblock \emph{\bibinfo{journal}{Physical review E}}
  \textbf{\bibinfo{volume}{51}}, \bibinfo{pages}{4282} (\bibinfo{year}{1995}).

\bibitem{burstedde295simulation}
\bibinfo{author}{Burstedde, C.}, \bibinfo{author}{Klauck, K.},
  \bibinfo{author}{Schadschneider, A.} \& \bibinfo{author}{Zittarz, J.}
\newblock \bibinfo{title}{Simulation of pedestrian dynamics using a
  2-dimensional cellular automaton, 2001}.
\newblock \emph{\bibinfo{journal}{Physica A}} \textbf{\bibinfo{volume}{295}},
  \bibinfo{pages}{507}.

\bibitem{helbing2000simulating}
\bibinfo{author}{Helbing, D.}, \bibinfo{author}{Farkas, I.} \&
  \bibinfo{author}{Vicsek, T.}
\newblock \bibinfo{title}{Simulating dynamical features of escape panic}.
\newblock \emph{\bibinfo{journal}{Nature}} \textbf{\bibinfo{volume}{407}},
  \bibinfo{pages}{487--490} (\bibinfo{year}{2000}).

\bibitem{steffen2010methods}
\bibinfo{author}{Steffen, B.} \& \bibinfo{author}{Seyfried, A.}
\newblock \bibinfo{title}{Methods for measuring pedestrian density, flow, speed
  and direction with minimal scatter}.
\newblock \emph{\bibinfo{journal}{Physica A: Statistical mechanics and its
  applications}} \textbf{\bibinfo{volume}{389}}, \bibinfo{pages}{1902--1910}
  (\bibinfo{year}{2010}).

\bibitem{li2015crowded}
\bibinfo{author}{Li, T.} \emph{et~al.}
\newblock \bibinfo{title}{Crowded scene analysis: A survey}.
\newblock \emph{\bibinfo{journal}{IEEE Transactions on Circuits and Systems for
  Video Technology}} \textbf{\bibinfo{volume}{25}}, \bibinfo{pages}{367--386}
  (\bibinfo{year}{2015}).

\bibitem{stauffer2000learning}
\bibinfo{author}{Stauffer, C.} \& \bibinfo{author}{Grimson, W. E.~L.}
\newblock \bibinfo{title}{Learning patterns of activity using real-time
  tracking}.
\newblock \emph{\bibinfo{journal}{IEEE Transactions on pattern analysis and
  machine intelligence}} \textbf{\bibinfo{volume}{22}},
  \bibinfo{pages}{747--757} (\bibinfo{year}{2000}).

\bibitem{viola2005detecting}
\bibinfo{author}{Viola, P.}, \bibinfo{author}{Jones, M.~J.} \&
  \bibinfo{author}{Snow, D.}
\newblock \bibinfo{title}{Detecting pedestrians using patterns of motion and
  appearance}.
\newblock \emph{\bibinfo{journal}{International Journal of Computer Vision}}
  \textbf{\bibinfo{volume}{63}}, \bibinfo{pages}{153--161}
  (\bibinfo{year}{2005}).

\bibitem{khan2016analyzing}
\bibinfo{author}{Khan, S.~D.}, \bibinfo{author}{Bandini, S.},
  \bibinfo{author}{Basalamah, S.} \& \bibinfo{author}{Vizzari, G.}
\newblock \bibinfo{title}{Analyzing crowd behavior in naturalistic conditions:
  Identifying sources and sinks and characterizing main flows}.
\newblock \emph{\bibinfo{journal}{Neurocomputing}}
  \textbf{\bibinfo{volume}{177}}, \bibinfo{pages}{543--563}
  (\bibinfo{year}{2016}).

\bibitem{liu2009extraction}
\bibinfo{author}{Liu, X.}, \bibinfo{author}{Song, W.} \&
  \bibinfo{author}{Zhang, J.}
\newblock \bibinfo{title}{Extraction and quantitative analysis of microscopic
  evacuation characteristics based on digital image processing}.
\newblock \emph{\bibinfo{journal}{Physica A: Statistical Mechanics and its
  Applications}} \textbf{\bibinfo{volume}{388}}, \bibinfo{pages}{2717--2726}
  (\bibinfo{year}{2009}).

\bibitem{hoogendoorn2005pedestrian}
\bibinfo{author}{Hoogendoorn, S.~P.} \& \bibinfo{author}{Daamen, W.}
\newblock \bibinfo{title}{Pedestrian behavior at bottlenecks}.
\newblock \emph{\bibinfo{journal}{Transportation science}}
  \textbf{\bibinfo{volume}{39}}, \bibinfo{pages}{147--159}
  (\bibinfo{year}{2005}).

\bibitem{vu15heads}
\bibinfo{author}{Vu, T.}, \bibinfo{author}{Osokin, A.} \&
  \bibinfo{author}{Laptev, I.}
\newblock \bibinfo{title}{Context-aware {CNNs} for person head detection}.
\newblock In \emph{\bibinfo{booktitle}{International Conference on Computer
  Vision ({ICCV})}} (\bibinfo{year}{2015}).

\bibitem{rodriguez11b}
\bibinfo{author}{Rodriguez, M.}, \bibinfo{author}{Sivic, J.},
  \bibinfo{author}{Laptev, I.} \& \bibinfo{author}{Audibert, J.-Y.}
\newblock \bibinfo{title}{Density-aware person detection and tracking in
  crowds}.

\end{thebibliography}

\end{document}